\documentclass{ecai}
\usepackage{caption}
\usepackage{graphicx}
\usepackage{latexsym}

 \usepackage{amsmath}
 \usepackage{amsfonts}
\usepackage{amsthm}
\usepackage{xcolor}

\usepackage{threeparttable}
\usepackage{multirow}
\usepackage{multicol}

\usepackage{makecell}
\usepackage{tabularborder}

\newtheorem*{remark}{Remark}

\usepackage[colorlinks=true, urlcolor=blue, linkcolor=red]{hyperref}

\newcommand{\mbf}[1]{\mathbf{#1}}
\newcommand{\mbb}[1]{\mathbb{#1}}
\newcommand{\mcal}[1]{\mathcal{#1}}

\newcommand{\citet}[1]{\cite{#1}}
\newcommand{\citep}[1]{\cite{#1}}

\def\L{\mathbf{L}}
\usepackage{amsthm}
\def\R{\mathbb{R}}

\def\x{\mathbf{x}}

\def\X{\mathbf{X}}

\def\R{\mathbb{R}}

\def\D{\mbf{D}}

\def\L{\mbf{L}}

\def\Y{\mathbf{Y}}

\def\A{\mbf{A}}
\def\B{\mbf{B}}
\def\S{\mbf{S}}

\paperid{426}        

\begin{document}

\begin{frontmatter}

\title{Graph Augmentation for Cross Graph Domain Generalization}

\author[A]{\fnms{Guanzi}~\snm{Chen}$^\dagger$\thanks{\textit{Email:guanzichen99@gmail.com; yangli@sz.tsinghua.edu.cn} }}
\author[A]{\fnms{Jiying Zhang}$^\dagger$}
\author[A]{\fnms{Yang Li} }
%


\address[A]{Tsinghua University}

\begin{abstract}
Cross-graph node classification, utilizing the abundant labeled nodes from one graph to help classify unlabeled nodes in another graph, can be viewed as a domain generalization problem of graph neural networks (GNNs) due to the structure shift commonly appearing among various graphs.
Nevertheless, current endeavors for cross-graph node classification mainly focus on model training. Data augmentation approaches, a simple and easy-to-implement domain generalization technique,  remain under-explored.
In this paper, we develop a new graph structure augmentation for the cross-graph domain generalization problem.
Specifically, low-weight edge-dropping is applied to remove potential noise edges that may hinder the generalization ability of GNNs, stimulating the GNNs to capture the essential invariant information underlying different structures. Meanwhile, clustering-based edge-adding is proposed to generate invariant structures based on the node features from the same distribution. Consequently, with these augmentation techniques, the GNNs can maintain the domain invariant structure information that can improve the generalization ability. The experiments on out-of-distribution citation network datasets verify our method achieves state-of-the-art performance among conventional augmentations.
\end{abstract}

\end{frontmatter}

\section{Introduction}
Recently, Graph Neural Networks (GNNs)~\cite{kipf2017semi} have become popular in performing machine learning tasks on graph-structured data, showing superior performance in various application domains, such as social network analysis~\cite{kipf2017semi},
recommendation systems~\cite{wu2022graph}, traffic prediction~\cite{jiang2022graph}, chemical molecules~\cite{yang2019analyzing,zhang2022fine}, and so on.
Despite the remarkable success, due to the heavy reliance on the I.I.D assumption that the training and testing data are independently drawn from an identical distribution\cite{pan2010survey}, GNNs are often brittle and susceptible to \textit{distribution shift}, which widely exists in the real-world scenarios. For example, social networks may differ when collected from different communities, molecules may have different structures, and transition networks change as time goes by. Graph domain generalization, aiming to improve the generalization performance under unseen distribution shifts has become a critical problem~\cite{li2022out}.

Due to the complexity of graph distribution shift type and the diversity of graph learning tasks, domain generalization on graphs have many settings, such as graph classification problem with structure shifts~\cite{chen2022invariance} and single graph node classification problem~\cite{yu2022finding}.
In this paper, we focus on the cross-graph node classification tasks with \textit{Out-Of-Distribution} (OOD) graph structures~\cite{wuhandling}.
Specifically, the graph model will  be trained firstly on several graphs in different domains  by node-level full supervised learning and directly predict the node labels of graphs from an unseen domain. Formally, suppose that the node feature distribution is $p(\mcal X)$, the node label distribution is $p(\mcal Y)$, and the graph structure distribution is $p(\mcal A| \mcal{E} )$, where $\mcal E$ is the environment variable and the cause of shift structure distribution can be considered to be
the environment difference.
Thus, the task can be formulated as follows.
\begin{align}
\label{eq:objective}
    \mathop{\min}_{f} \mathop{\max}_{\varDelta } \mbb{E}_{(\X,Y)\sim p(\mcal X,\mcal Y);\A \sim p(\mcal A| \mcal{E} = \varDelta)} \sum_{v\in V} l(f(G(\X, \A))_v, Y_v),
\end{align}
where  $p(\mcal X,\mcal Y)=p(\mcal X)p(\mcal Y | \mcal X)$, $f$ is a GNN with graph $G$ as input for predicting the class of nodes, and $Y_v$ represents the label of node $v$. $l$ denotes the classification loss function, such as cross-entropy.
The optimal $f$ that is able to well process the graph from new environment can be obtained via  optimizing  the objective Eq.~\eqref{eq:objective}. The task widely exists in the real world~\cite{wuhandling}.
For instance, in citation networks, the node features are in the same distribution, but each network formed by different relationships has its own structure distribution, and the model can be trained on  several citation networks to improve the domain generalization ability. 

In general, there exist three perspectives to handle domain generalization problems on graph data~\cite{li2022out}, including  model-level methods~\cite{li2021disentangled,li2022ood}, model-training-level methods~\cite{liu2022confidence,wu2022discovering} and data-level methods~\cite{rongdropedge,you2020graph}.  For cross-graph node classification with OOD structure, 
from the model-training perspective, \citet{wuhandling} 
proposed an invariant learning method EERM, which can be viewed as the extension of ERM~\cite{arjovsky2020invariant}. They focus on designing a new loss involving risk variance minimization to improve the generalization of the learned GNN. On the other hand, 
 data-level approaches like data augmentation, a simple and
easy-to-implement domain generalization technique, remains under-explored.

In this paper, from the data-level standpoint, we propose a new simple and effective graph augmentation  for node-level tasks with OOD structure. Our goal is to utilize data augmentation to expand the volume and diversity of training samples, and simultaneously enhance the ability of models to acquire invariant information between different environments. 
Based on the fact that the distribution shift is caused by the structure,  we first utilize an edge sampling method to boost the volume and variety of the training data. The edge sampling is implemented by a low-weight edge-dropping technique that will drop out some edges that are potential noise for the OOD tasks, leaving a subset of the edge that represents the essential structure of the original graph. 

\begin{figure}[th]
    \centering
    \includegraphics[width=0.88\linewidth]{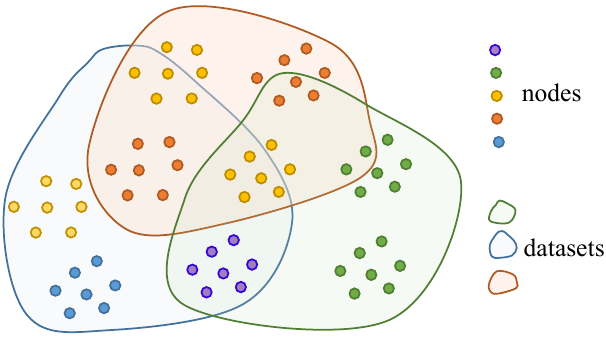}
     \caption{Example of OOD structure. The small circles are utilized to indicate the node features in the feature space. The three big circles mean three datasets with different node sets. Each of them can construct its own graph structure. Three node sets can form three graphs with OOD structures  }
    \label{fig:node_space}
\end{figure}

Second, we need to explore the domain invariant information among the various structures.  Due to the same distribution of nodes (i.e. $p(\mcal X)$) in the structure OOD task, it is possible to use the Identity Distribution node attributes to capture the invariant structure information across graphs. Normally, the graph structure is commonly applied to reflect the relationship between nodes. Two nodes that are connected usually imply they are similar in the homogeneous graph, in other words, they are closed in the feature space. Hence, one promising augmentation operation is to generate new edges that mark the high similarity of connected nodes with initial features. Because the node features are sampled from the same feature space (Figure~\ref{fig:node_space} ), the added edge reflects the essential invariant topology of the node features, thereby amplifying the  capacity of the learned GNN's ability to perceive the prototypes underlying the node features. In addition, the generated essential topology may not focus on the domain-specific information, on the contrary, it maintains the domain-invariant information that can improve the generalization ability of GNN. 
The experiments
conducted on citation network datasets show that the proposed approach achieves the best performance
among all baselines, validating the effectiveness and generalization ability of our method.

In summary, the contributions of this paper can be highlighted as follows:
   (1) We propose a new edge-dropping method to remove the potential noise edges while retaining the key structure of the initial graph.
(2) A spectral clustering-based edge-adding strategy is developed to boost the learned model to perceive the global node clusters in the feature space, which is also the invariant information in the structure OOD datasets.
(3) The extensive experiments demonstrate the proposed augmentation achieves competitive performance in the cross-graph node classification and state-of-the-art performance among typical graph augmentations.

\section{Related work}
\paragraph{Domain Generalization on Graph.}
Recently, the research of graph domain generalization (DG) to improve the generalization ability of Graph Neural Networks under distribution shifts has come into the spotlight.
Mostly, the domain generalization methods can be divided into three categories\cite{li2022out}:
model-level methods~\cite{li2021disentangled,li2022ood}, model-training-level methods~\cite{liu2022confidence,wu2022discovering} and data-level methods~\cite{rongdropedge,you2020graph}. Among them, the data-level methods are attractive  due to their simplicity and ease of deployment. The most important data-level method is data augmentation.
However, the existing data augmentations for DG are mainly specifically designed for inductive graph classification which is a graph-level task~\cite{kong2020flag}, or single graph node classification in which the distribution shift happens in node feature~\cite{park2021metropolis}, without considering the inductive cross graph node classification that the distribution shift occurs in graph structures. Therefore, it remains open to using data augmentation to improve the model generalization ability in cross-graph node classification.

\paragraph{Structure Modification.}
Existing methods involving edge modification tend to tailor for i.i.d. data. Those methods aim at dropping out the potentially
task-irrelevant edges from input graphs. For example, \citet{luo2021learning} proposed a topology denoising algorithm for filtering out the task-irrelevant edges. \citet{zhao2021data} raised a 
 learnable augmenter to generate and remove the edges in the graph augmentation, where the augmenter is  optimized by the downstream task, capturing the information that is highly related to 
the environment. \citet{zheng2020robust} present a graph sparsification approach to remove potentially task-irrelevant edges via deep neural networks. Despite those methods being successful in the semi-supervised node classification tasks,
it is inappropriate to make the modified graph structure only concentrate on the domain-related task in the cross-graph domain generalization. 
These highly domain-related techniques  would focus so much on information about the current domain that they ignore domain-invariant information which is critical for domain generalization.

\paragraph{Cross Graph Node Classification.}
Cross-graph node classification is a common task in practice. For example, in protein-protein interaction networks, one can leverage the abundant functional information from a source network to help predict the functionalities of proteins in a newly formed target network.
Many methods~\cite{dai2022graph,shen2020adversarial,yang2022robust} propose to address the cross-graph node classification problem by integrating graph neural networks and adversarial domain adaptation, in which the graph neural network is used as a feature extractor to learn discriminative node representation while adversarial learning is utilized to capture domain invariant node representations. AdaGCN\cite{dai2022graph} employs GCN\cite{kipf2017semi} to preserve node attribute and topological structure and uses WDGRL~\cite{shen2018wasserstein} as the adversarial learning strategy. ACDNE\cite{shen2020adversarial} design a deep network embedding module with two feature extractors following the DANN~\cite{ganin2016domain} adversarial learning strategy. RGDAL\cite{yang2022robust} inherits the framework and further filter noisy information via constrained graph mutual information. Nevertheless, they have not considered the cross-graph node classification task under the domain generalization context, in which the target graphs cannot be accessed during training.

s

\section{Methodology}

\subsection{Preliminaries}
\paragraph{Notations.}
Let $G(V,E,\A,\X)$ denote a graph with node set $V$, edge set $E$, adjacency matrix $\A \in \R^{|V|\times |V|}$ and node feature matrix $\X = \{\x_1;\x_2; \cdots,\x_{|V|}\} \in\R^{|V|\times d}$. We denote the Hadamard product by $\odot$. Besides, if not specified, we use boldface letter $\x\in \R^n$
 to indicate an $n$-dimensional vector, where $x_i$ is
the $i^\text{th}$ entry of $\x$. We use a boldface capital letter
$\mbf A\in\R^{m\times n}$ to denote an $m$ by $n$ matrix and use $A_{ij}$ to denote its ${ij}^\text{th}$ entry. 
\paragraph{Spectral Clustering.}
Spectral clustering in multivariate statistics involves using the eigenvectors of a matrix based on the similarity matrix (affinity matrix) to cluster data points. The technique conventionally requires computing the $k$ smallest eigenvalues and corresponding eigenvectors of the affinity matrix.
The eigenvectors are then used to form a new lower-dimensional space in which \textit{k-means} clustering or another clustering algorithm can be applied to separate the data points into clusters.

More specifically, the process of spectral clustering involves finding the $k$ smallest eigenvalues and corresponding eigenvectors of the normalized Laplacian matrix $\L_s = \D_s^{-1/2} \mbf S \D_s^{-1/2}$, where $\S$ is the affinity matrix (similarity matrix) and $\D_s$ is the diagonal matrix with element $\D_s(i,i)=\sum_j \S_{ij}$.
The the $k$ eigenvectors will be Concatenated together to form a matrix $\mbf U' = [\mathbf{u}_1, \mathbf{u}_2, \ldots, \mathbf{u}_k]$. Ultimately, the rows of $\mbf U'$, in which each row represents a data point, will be clustered by a specific clustering algorithm such as \textit{k-means}.

\subsection{Graph Augmentation for Cross Graph DG}
In order to make the graph neural  network able to perceive the essential distribution of node features in the feature space, we propose a two-step strategy. First, remove the minor edges and leave the major ones. Intuitively, the removed edges contain potential noise that hinders the GNN  from capturing the node feature distribution while the left edges should mainly maintain the key structure of the graph that embeds the essential information of the data.

Second, add some edges by clustering. The remaining edges in the first step may  only be the partial edges of the key topology, which would be insufficient to reflect the node feature distribution.  Therefore, adding some crucial edges that can reflect the node feature distribution is a natural operation to improve invariant information learning in GNN.

Actually, the second step can be canceled if the edges remaining in the first step is good enough for GNN to perceive the essential information underlying the node features.

\subsubsection{Low-Weights Edge Dropping} \label{subsec:low-weights}
This part proposes an edge-sampling strategy for sampling the core graph structure. The edge sampling is implemented by a low-weight edge-dropping operation. Intuitively, when dropping edges from a graph, we can use a Bernoulli distribution to determine whether each edge exists or not, based on some probability parameter $\rho$. For the whole graph, we need a compute a weight matrix to assign the sampling probability for each edge. 

\begin{figure*}[bht]
    \centering
    \begin{minipage}[s]{0.4\linewidth}
    \centering
    \includegraphics[width=0.9\linewidth]{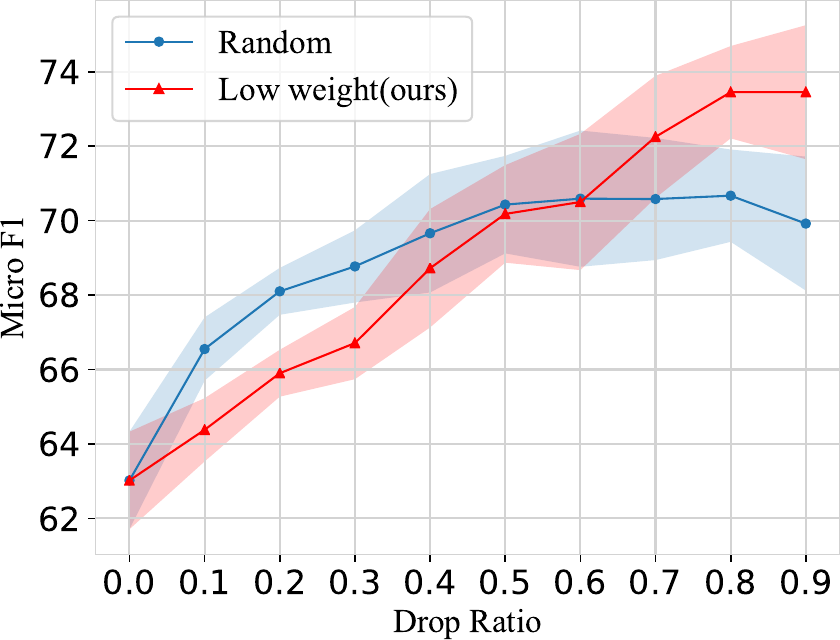}
    \end{minipage}
\quad
    \begin{minipage}[s]{0.4\linewidth}
    \centering
    \includegraphics[width=0.9\linewidth]{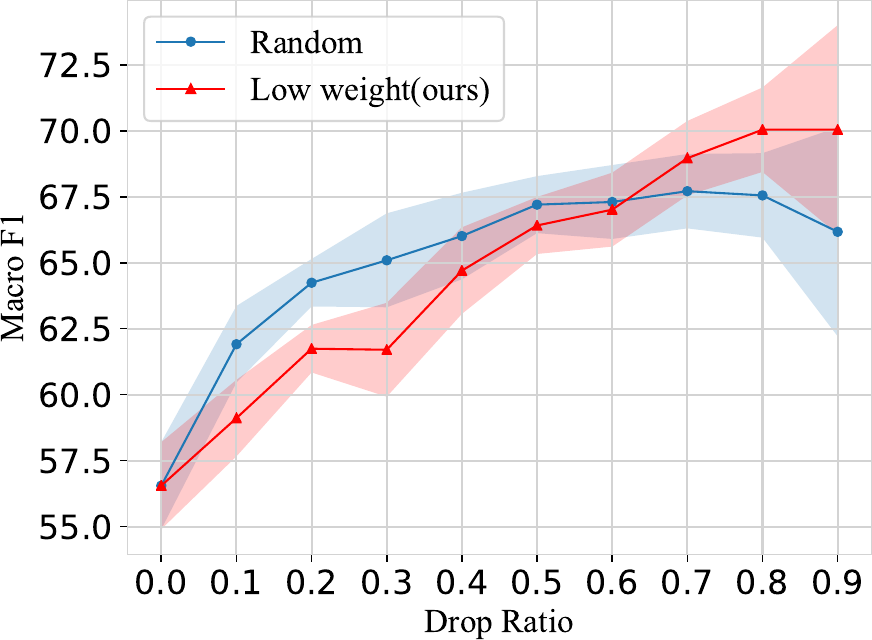}
    \end{minipage}

     \caption{The comparison between random edge-dropping and low weight dropping.(backbone: GIN, AD $\rightarrow $ C)}
    \label{fig:all_acc_gin}
\end{figure*}

\paragraph{Edge Weights Computation.}
The graph Laplacian is commonly used to represent graphs and drive the graph spectral convolution neural networks. 
The elements in the graph Laplacian matrix indicate the connection relations between the vertices in the corresponding graph. 
Specifically,
the Laplacian matrix denote as 
$\L  = \mbf I-\mathbf{D}^{-\frac{1}{2}}\mathbf{A}\mathbf{D}^{-\frac{1}{2}}$, which can be rewritten as:
\begin{equation}
\L = \mbf I- \A \odot
 \begin{bmatrix} \frac{1}{\sqrt{d_1}\sqrt{d_1}} & \frac{1}{\sqrt{d_1}\sqrt{d_2}} & \cdots & \frac{1}{\sqrt{d_1}\sqrt{d_n}} \\ \frac{1}{\sqrt{d_2}\sqrt{d_1}} & \frac{1}{\sqrt{d_2}\sqrt{d_2}} & \cdots & \frac{1}{\sqrt{d_2}\sqrt{d_n}}\\
\ \vdots & \vdots & \ddots & \vdots 
\\ \frac{1}{\sqrt{d_n}\sqrt{d_1}} & \frac{1}{\sqrt{d_n}\sqrt{d_2}} & \cdots & \frac{1}{\sqrt{d_n}\sqrt{d_n}} \end{bmatrix}.
\end{equation}
The graph Laplacian can be understood from the Total Variation of the graph signal. Given the signal $x\in \R^{|V|}$ on graph $G$, its total variation can be expressed as follows:  

\begin{equation}
\label{eq:laplacian_total_variance}
\begin{aligned}
    x^{\top} \L x&=x^{\top}  x-x^{\top} \mathbf{D}^{-\frac{1}{2}}\mathbf{A}\mathbf{D}^{-\frac{1}{2}} x 
    =\sum \limits _{i=1}^{|V|} x_{i}^{2}-\sum\limits_{i=1}^{|V|} \sum\limits_{j=1}^{|V|} \frac{1}{\sqrt{d_{i}d_j}}  x_{i} x_{j} \\
    &=\frac{1}{2}\left(\sum\limits_{i=1}^{|V|}  x_{i}^{2} -2 \sum\limits_{i=1}^{|V|} \sum\limits_{j=1}^{|V|} \frac{1}{\sqrt{d_{i} d_j}}  x_{i} x_{j}+\sum\limits_{j=1}^{|V|} x_{j}^{2}\right) \\
    &=\frac{1}{2} \sum\limits_{i=1}^{|V|} \sum\limits_{j=1}^{|V|}\left(\frac{x_{i}}{\sqrt{d_{i}}}-\frac{x_{j}}{\sqrt{d_{j}}}\right)^{2}.
\end{aligned}
\end{equation}

In particular, we consider using the Laplacian matrix to determine which edges should be removed from the original graph. 
From  the last line
 in Eq.~\eqref{eq:laplacian_total_variance}, we know the $\frac{1}{\sqrt{d_i}}$ can be viewed as the importance of the signal on node $i$. Assume that the edge connecting two significant nodes is more vital for representing the key structure of the graph. Thus the product of $\frac{1}{\sqrt{d_i}}$ and $\frac{1}{\sqrt{d_j}}$ can be used to measure the criticality of edge $e_{ij}$. Formally, the edge weight matrix is:
 \begin{align}
     \mbf P=\mathbf{D}^{-\frac{1}{2}}\mathbf{A}\mathbf{D}^{-\frac{1}{2}},
 \end{align}
 with elements $P_{ij} = \frac{1}{\sqrt{d_id_j}}$ if $A_{ij}>0$, otherwise $0$. The intuition behind it is that the edge is  crucial if its incident nodes have a low degree. In other words, the edges connected to low-degree nodes are more likely to affect the properties of the graph, such as connectivity. This is because, if a graph is disconnected, these low-degree nodes are often the first to be disconnected and thus can be an important indicator of the graph's connectivity.
 In addition, from the GCN perspective, the proposed edge weight matrix $\mbf P$ is actually the message aggregation matrix, thereby elements with relatively large values  in $\mbf P$  indicate the key process of message passing. Remaining the key message passing paths and removing the minor paths are likely to avoid aggregating noise that may hurt the generalization of GNNs.

\paragraph{Adaptive Sampling Strategies.}
There exist many approaches that can be used to construct sampling strategies \cite{gao2021training} based on the edge weights matrix $\mbf P$. 
The principle of the sampling strategies is the larger-weight edges have a higher probability to remain because those undropped edges represent the key structure of the original graph.
\begin{itemize}
    \item \textbf{Threshold Cutoff:} We determine a threshold $\tau$ for edge weights based on their distribution and use it to decide whether to sample each graph edge with the default probability $\rho$ or to preserve it as an undropped critical edge. In particular, edges with weights below $\tau$ are sampled using the default probability, while those with weights above $\tau$  are critical and preserved to maintain the graph key structure. This can be formulated as:
    \begin{align}
    \label{eq:threshold_cutoff_dorpedge}
        P^{D}_{ij}=\left\{\begin{array}{cl}
	\rho, & \text{if } P_{ij}<\tau \text{ and } (v_i,v_j)\in E, \\
    1, & \text{if } P_{ij}\ge\tau \text{ and } (v_i,v_j)\in E, \\
	0, & \text{otherwise.} 
        \end{array}\right.
    \end{align}    
 
In practice, in order to control the edge dropping rate, we use the $k$th-smallest edge weight to determine the threshold $\tau$, i.e. $\tau = f_{min}(\mbf e^{D}, \alpha |E|)$, where $f_{min}(\cdot,k)$ is the $k$th-smallest value function, $\mbf e^{D} \in \R^{|E|}$ is the edge weight vector unfolded from $\mbf P^D$ and the $\alpha \in [0,1]$ is the edge dropping ratio. 
\item \textbf{Division Normalization:}
We employed a division function to normalize the edge weights to a range of $[0, 1]$, and subsequently used the normalized weights to determine the sampling matrix $\mbf P^D$. To calculate the probability of edge sampling given an edge weight of $\omega$ and a function parameter of $\tau$, we can apply the formulation $P^D_{ij} = 1 - (1 - P_{ij}) * \tau/(\tau + \omega)$.

\item \textbf{CDF Normalization:}
Our approach involves the normalization of edge weights, using the cumulative distribution function (CDF), which transforms them to a range of $[0, 1]$. Next, we determine the sampling matrix $\mbf P^D$ using the normalized weights. By utilizing our method's equation that requires the edge weight $\omega$ and its corresponding CDF $f(\omega)$, we can compute the probability of edge sampling, represented as $P^D_{ij} = P_{ij} + (1 - P_{ij}) * f(\omega)$.
    
\end{itemize}

 \paragraph{Low-weights Edge Dropping Augmentation.} 
The  $\mbf P^{D}$ can be used as the probability matrix in the Bernoulli distribution for generating the binary matrix. Specifically,  the augmenter is represented as:
\begin{align}
    \A' &= \A \odot \mbf M, \\
    \mbf M & \sim  \text{Bernoulli}(\mbf P^D),
\end{align}
where $M_{ij} \sim \text{Bernoulli}(P^D_{ij})$.
Eventually, the new graph can be led by the low-weight edge-dropping augmentation:
\begin{align}
G^{D}(V, E^D, \mbf A^D, \mbf X) = G(V, E, \A', \mbf X).
\end{align}
The augmenter would generate different augmentation $G^{D{(t)}}$ at each epoch $t$, which increases the diversity of the training data.

Although the low-weight edge dropping remove the potential noise edge that may hurt the generalization of the learned GNNs, the remaining edges are not always adequate to reflect the whole invariant structure that benefits the OOD task. Inspired by this, we consider using a clustering-based method to add some edges into  $G^{D}$ for improving the GNN perceive the global node feature distribution (clustering prior), which is the invariant information on node-level tasks with OOD structure.

 \begin{table*}[ht]
\def\p{$\pm$} 
\centering
\setlength\tabcolsep{4 pt}
\caption{The prediction accuracy on the Citation benchmark (A: ACM, D: DBLP, C: Citation)}
\scalebox{0.883}{
\begin{tabular}{c|cc|cc|cc}
\toprule 
\multirow{2}{*}{Methods} & \multicolumn{2}{c|}{AD $\to$ C}    & \multicolumn{2}{c|}{AC $\to$ D}  & \multicolumn{2}{c}{CD $\to$ A} \\
 & Micro F1 &  Macro F1 & Micro F1 & Macro F1  & Micro F1 & Macro F1 \\
\midrule
GCN~\cite{kipf2017semi} & 63.64 $\pm$ 0.64 & 61.16 $\pm$ 0.80 &   64.86 $\pm$ 0.42 & 61.56 $\pm$ 0.46 &  59.77 $\pm$ 0.53 & 57.78 $\pm$ 0.81  \\                 
GIN~\cite{xu2018powerful} &  63.02 $\pm$ 1.31 & 56.55 $\pm$ 1.65  & 63.73 $\pm$ 1.31 & 58.63 $\pm$ 2.41   &56.45 $\pm$ 1.45 & 53.55 $\pm$ 2.64   \\                       
GAT~\cite{velivckovicgraph} &   61.44 $\pm$ 1.07 & 57.93 $\pm$ 0.77  &  64.26 $\pm$ 0.94 & 60.80 $\pm$ 0.94   &  59.11 $\pm$ 0.83 & 56.21 $\pm$ 0.89 \\                       
SGC~\cite{wu2019simplifying} &  66.12 $\pm$ 1.07 & 61.93 $\pm$ 1.22   &  67.15 $\pm$ 0.75 & 63.42 $\pm$ 1.02  &  61.92 $\pm$ 0.92 & 57.06 $\pm$ 1.79   \\                                                
\hline                                         
GCN+(Ours) & \underline{71.62 $\pm$ 0.75}$\uparrow_{7.98}$ & \underline{66.09 $\pm$ 0.56}$\uparrow_{4.93}$ &    \underline{71.45 $\pm$ 0.55}$\uparrow_{6.59}$ & \underline{67.19 $\pm$ 0.72}$\uparrow_{5.63}$  &   \underline{66.44 $\pm$ 0.52}$\uparrow_{6.67}$ &  \underline{60.70 $\pm$ 0.70}$\uparrow_{2.92}$      \\
GIN+(Ours) & \textbf{74.77 $\pm$ 0.60}$\uparrow_{11.75}$ & \textbf{71.79 $\pm$ 1.18}$\uparrow_{15.24}$ &  \textbf{72.55 $\pm$ 0.60} $\uparrow_{8.82}$ & \textbf{70.05 $\pm$ 1.71}$\uparrow_{11.42}$    &   \textbf{69.43 $\pm$ 0.65} $\uparrow_{12.98}$ & \textbf{68.77 $\pm$ 1.07}$\uparrow_{15.22}$        \\
\bottomrule
    \end{tabular}
    }
    \label{tab:citation_benchmark}    
\end{table*}
\subsubsection{Spectral Clustering Based Edge Adding}
As we all know, the node in the feature space has its own position. The nodes with similar features are usually closed in the feature space. In the node-level tasks with OOD structure, the node feature distribution is invariant (c.f. Figure \ref{fig:node_space})..
We consider generating the invariant structure from node features. A simple and effective approach is to  employ a clustering method to get the essential structure of the homogeneous graph. Using the cluster graph as the training data will promote the  capacity of the GNN to perceive the prototypes underlying the node features, thereby enhancing the learned GNN generalized ability.

\paragraph{Cluster Induced Graph Generation.}
Here we apply spectral clustering.
Let the number of clusters be $\mcal K >1$. 
By spectral clustering of node features, we can obtain which cluster each node belongs to. For each cluster, we assign it a pseudo label $k ~(k=1,2,3,...,\mcal K)$.   Let the pseudo label of node $v_i$  denote as $\mcal C(v_i)$.
After getting the cluster label of each node, the nodes with  the same label can connect to each other, that is to say, each cluster will be transformed into a complete subgraph. 
We first use a matrix $\mbf B\in\R^{|V|\times \mcal K}$ to formulate the cluster information:
\begin{align}
\label{eq:incidence_matrix}
    B_{ij} = \left\{\begin{array}{cl}
	1, &\text{if } \mcal C(v_i) = j, \\
	0, & \text{otherwise.}
 \end{array}\right.
\end{align}
Then the adjacency matrix of the cluster-induced graph $G^G(V,E^G,\A^G,\X)$ can be expressed as:
\begin{align}
    \A^G = \operatorname{sign}(\mathbf B \mbf B^{\top}),
\end{align}
where $sign(\cdot)$ signifies the symbolic function. A simple way to get the edge from $\A^G$ is randomly sampling edges in $G^G$. 

Alternatively, we can assign a weight matrix of edges to $G^G$ so that the sampling can be based on a meaningful probability, thus improving the interpretation of the edge-adding operation.

\paragraph{Edge Weights Computation for Edges Sampling.}
We can generate the edge weights based on the $\B$ in Eq.~\eqref{eq:incidence_matrix}.  Actually, the element in $\mbf B\mathbf{B}^{\top}$ is $\sum_k B_{ik}B_{jk}$. We can consider involving the number of nodes in each cluster. Assume that the fewer the number of nodes contained in a cluster, the more important the edges formed by the cluster. This is because those small clusters contain more critical topology information, such as connectivity. On the other hand, the big clusters are easily sampled due to the dense connections in the complete subgraph. Thus, define the number of nodes in cluster $j$ as  $\delta_j:= \sum_{i=1}^{\mcal |V|} B_{ij}$ and add it into the edge weight computation, we have 
$P^{G}_{ij} = \sum_k \frac{1}{\delta_k} B_{ik}B_{jk}$, which can be rewritten as matrix form $\mathbf B \mbf D_e^{-1} \mbf B^{\top}$. Formally, the edge weights in $G^G$ can be calculated as follows:
\begin{align}
    \mbf P^{G} = \mathbf B \mbf D_e^{-1} \mbf B^{\top},
\end{align}
where  
$\D_e\in\R^{|\mcal K|\times |\mcal K|}$ is a diagonal matrix with element  $D_e{(j,j)} =\delta_j$. 
The $\mbf P^{G}$ can be viewed as the transition matrix in the hypergraph random walk~\cite{zhang2022hypergraph} (each cluster is a hypergraph edge and $\mbf B$ is the incidence matrix) and the higher transition probability between two nodes means the edge connecting the nodes is more important.   
Next, based on the edge weights $\mbf P^{G}$, we can use the  sampling strategies as subsection ~\ref{subsec:low-weights}  to sample the edges. Taking \textit{threshold cutoff} as an example, given the probability $\epsilon, \rho$ where $\epsilon<\rho$, the sampling probability is:
\begin{align}
\label{eq:threshold_cutoff_dropadd}
    P^{G}_{ij}=\left\{\begin{array}{cl}
	\epsilon, & \text{if } P^G_{ij}<\tau \text{ and } \A^G_{ij} =1, \\
    \rho, & \text{if} P^G_{ij}\ge\tau \text{ and } \A^G_{ij}=1, \\
	0, & \text{otherwise.} 
        \end{array}\right.
\end{align}

\begin{align}
\label{eq:threshold_cutoff_dropad}
    P^{G}_{ij}=\left\{\begin{array}{cl}
    \rho, & \text{if } P^G_{ij}<\tau \text{ and } \A^G_{ij}=1, \\
	1, & \text{otherwise.} 
        \end{array}\right.
\end{align}
where the threshold $\tau$ is determined by the $k$th-largest value, which can be implemented by $k$th-smallest value function, i.e.
$\tau = -f_{min}(-\mbf e^{G}, \beta |E^G|)$, where $f_{min}(\cdot,k)$ is the $k$th-smallest value function, $\mbf e^{G} \in \R^{|E^G|}$ is the edge weight vector and the $\beta \in [0,1]$ is the edge adding ratio.

The  $\mbf P^{G}$ can be used as the probability matrix in the Bernoulli distribution for generating the binary matrix. Specifically,  the augmenter is represented as:
\begin{align}
    \A'' &= \A^G \odot \mbf M,\\
    \mbf M  \sim & \text{Bernoulli}(\mbf P^G),
\end{align}
where $M_{ij} \sim \text{Bernoulli}(P^{G}_{ij})$.
Ultimately, the amended cluster-induced graph is:
\begin{align}
G^{C}(V^C, E^C, \mbf A^C, \mbf X^C) = G(V, E^G, \A'', \mbf X).
\end{align}
Note that the $G^{C}$ is a separate graph augmentation that can be directly applied to the domain generalization tasks.

\paragraph{Edge Adding Augmentation.}

A simple way to add edges to $G^D$ is to select edges from $\A^C$ and merge them directly into $\A^D$, i.e.
\begin{align}
\label{eq:simple_new_graph}
    \A^{N} = \operatorname{sign}(\A^C + \A^D).
\end{align}
Alternatively, we also can use other mix-up~\cite{han2022g} methods to export a mix-up augmenter, formally,
\begin{align}
    \A^{N} = \operatorname{mix-up}(\A^C, \A^D),
\end{align}
where $\operatorname{mix-up}$ is mix up function. For example, if $\operatorname{mix-up}$  is a weighted sum function, we have $\A^{N} = \eta \A^C + (1-\eta) \A^D $.

\begin{remark}
Note that our method is a computationally efficient method. In practice, spectral clustering, edge weights matrix, as well as edge weights sorting in \textit{threshold cutoff} can be calculated before training. So our approach does not impose any additional computational burden during training and testing.
\end{remark}

\subsection{Overall Loss.} 
Given the source graphs $\{(G^S_i, \mbf Y_i)\}_{i=1}^{D}$,
and one target graph $(G^T, \Y)$ with different structure distribution from source graphs
, the optimized objective of the node classification with OOD structure is: 
\begin{equation}
\begin{aligned}
    \mathop{\arg\min}_{f} & \mcal L(f,\{G^S_i, \mbf Y_i\}_{i=1}^{D}) \\
    &= \mathop{\arg\min}_{f} \frac{1}{D}\sum_{i=1}^D \sum_{v\in G_i^S}  l(f(T(G^S_i))_v, \mbf Y_{iv}),
\end{aligned}
\end{equation}
where $T(\cdot)$ represents the augmenter, $f(\cdot)$ is a GNN that predicts the node class, and $l(\cdot)$ is the Cross-Entropy loss for classification or Binary Cross-Entropy loss for multi-class classification.

\section{Experiment}

 \begin{table*}[ht]
\def\p{$\pm$} 
\centering
\setlength\tabcolsep{12 pt}
\caption{The comparison with different graph augmentations. The backbone is GIN. ( A: ACM, D: DBLP, C: Citation)}
\scalebox{0.95}{
\begin{tabular}{c|c|ccc}
\toprule 
Methods & Augmenters  &  Micro F1 &  Macro F1 & Acc  \\
\midrule
\multirow{7}{*}{AD $\to$ C} &  baseline & 63.02 $\pm$ 1.31 & 56.55 $\pm$ 1.65 & 86.35 $\pm$ 0.62                  
 \\                          
& edge dropping     & {70.67 $\pm$ 1.24} & 67.55 $\pm$ 1.60 & 88.46 $\pm$ 0.43         
  \\                         
 & edge adding   & 69.04 $\pm$ 0.65 & 65.32 $\pm$ 0.78 & 88.88 $\pm$ 0.24                
 \\
& feature masking   & 72.17 $\pm$ 0.76 & 68.11 $\pm$ 1.17 & \underline{89.66 $\pm$ 0.23}          
  \\                         
 & feature dropout  & 70.14 $\pm$ 1.60 & 63.20 $\pm$ 2.54 & 89.69 $\pm$ 0.43         
  \\                         
&  Low weight dropping (ours) &  \underline{73.45 $\pm$ 0.89} & \underline{70.04 $\pm$ 1.27} & 89.23 $\pm$ 0.44 
  \\                      
  & low weight + clustering (ours) & \textbf{74.77 $\pm$ 0.60} $\uparrow_{11.75}$ & \textbf{71.79 $\pm$ 1.18} $\uparrow_{15.24}$ & \textbf{89.77 $\pm$ 0.30} $\uparrow_{3.42}$ \\
  
  \hline
  \multirow{7}{*}{AC $\to$ D}                       
&  baseline   & 63.73 $\pm$ 1.31 & 58.63 $\pm$ 2.41 & 86.54 $\pm$ 0.41                                 
 \\                                            
& edge dropping & 71.19 $\pm$ 0.99 & 68.28 $\pm$ 1.01 & 88.66 $\pm$ 0.51                       
  \\                                           
 & edge adding  & 67.83 $\pm$ 0.49 & 64.71 $\pm$ 1.04 & 88.53 $\pm$ 0.18                          
 \\                                            
& feature masking & 71.31 $\pm$ 0.37 & 68.28 $\pm$ 0.65 & \textbf{89.46 $\pm$ 0.12}                            
  \\                                           
 & feature dropout &     70.11 $\pm$ 0.98 & 65.88 $\pm$ 1.88 & \underline{89.43 $\pm$ 0.27}                       
  \\   

&  Low weight dropping (ours) & \textbf{ 72.55 $\pm$ 0.60} & \underline{69.61 $\pm$ 1.22} & 88.54 $\pm$ 0.30  \\
&  Low weight + clustering (ours)     & \underline{72.46 $\pm$ 0.78} $\uparrow_{8.73}$ & \textbf{70.05 $\pm$ 1.71} $\uparrow_{11.42}$ & 88.52 $\pm$ 0.41 $\uparrow_{1.98}$     
  \\  
    \hline
  \multirow{7}{*}{CD $\to$ A}                       
&  baseline   &   56.45 $\pm$ 1.45 & 53.55 $\pm$ 2.64 & 83.04 $\pm$ 0.94                               
 \\                                            
& edge dropping &      64.72 $\pm$ 1.95 & 63.91 $\pm$ 2.22 & 85.63 $\pm$ 0.77                  
  \\                                           
 & edge adding  &  61.54 $\pm$ 0.73 & 60.86 $\pm$ 0.84 & 86.14 $\pm$ 0.20                        
 \\                                            
& feature masking &  62.28 $\pm$ 1.99 & 60.47 $\pm$ 2.64 & 85.95 $\pm$ 0.61                          
  \\                                           
 & feature dropout &    61.53 $\pm$ 1.04 & 59.64 $\pm$ 1.44 & 85.90 $\pm$ 0.38                        
  \\                                           
&  Low weight dropping (ours) &      \underline{69.08 $\pm$ 0.82} & \underline{68.10 $\pm$ 1.28} & \underline{87.22 $\pm$ 0.49}           
  \\  
& Low weight + clustering (ours) & \textbf{69.43 $\pm$ 0.65} $\uparrow_{12.98}$ & \textbf{68.77 $\pm$ 1.07} $\uparrow_{15.22}$ & \textbf{87.36 $\pm$ 0.34} $\uparrow_{4.32}$\\
\bottomrule
    \end{tabular}
    }
    \label{tab:tab:augmentation_GIN_all}    
\end{table*}
We organized the experiments on full-supervised node classification tasks with structure OOD to answer the following questions: 1) Does the proposed augmentation technique work within the SOTA GNN frameworks? 2) How does the proposed augmentation technique compare with typical graph augmentation methods regarding accuracy? 3) How does the proposed low-weight edge-dropping compare with random edge-dropping methods under different dropping rates? 4) How do different components impact the overall performance?

\begin{table}
\centering
\caption{Real-world graph datasets used in domain generalization}
\renewcommand\tabcolsep{5.0pt} 
\vspace{-0mm}
\scalebox{1}{
\begin{tabular}{l|c|c|c|c}
\toprule
Dataset& \# Nodes & \# Edges & \# Features & \# Classes  \\
\midrule
Citation
& 8935         & 15113         &6775   & 5     \\

DBLP
& 5484        & 8130        & 6775       & 5     \\

ACM
& 9360        & 15602         & 6775        & 5      \\

\bottomrule
\end{tabular}
}
\label{tab:dataset_citation}

\end{table}

 \begin{table*}[htb]
\def\p{$\pm$} 
\centering
\vspace{-3mm}
\setlength\tabcolsep{12 pt}
\caption{The comparison with different graph augmentations ( CD $\to$ A, A: ACM, D: DBLP, C: Citation)}
\scalebox{0.9}{
\begin{tabular}{c|c|ccc}
\toprule 
Methods & Augmenters  &  Micro F1 &  Macro F1 & Acc  \\
\midrule
\multirow{5}{*}{GIN}                     
&  w/o both (baseline)   &   56.45 $\pm$ 1.45 & 53.55 $\pm$ 2.64 & 83.04 $\pm$ 0.94    
 \\   
& only clustering edge  &   \underline{69.23 $\pm$ 0.72} & 68.37 $\pm$ 0.99 & 86.80 $\pm$ 0.39     \\ 
&  w/o clustering edge &      69.08 $\pm$ 0.82 & \underline{68.10 $\pm$ 1.28} & \underline{87.22 $\pm$ 0.49 }         
  \\                        
 & w/o low weight dropping & 56.89 $\pm$ 3.33 &  52.50 $\pm$ 3.34 &  87.08 $\pm$ 0.63
\\
& low weigh + clustering (ours) & \textbf{69.43 $\pm$ 0.65} & \textbf{68.77 $\pm$ 1.07} & \textbf{87.36 $\pm$ 0.34}
\\
\bottomrule
    \end{tabular}
    }
    \label{tab:ab_study_edge}    
\end{table*}

\subsection{Experiment Settings}
In the experiment, we focus on node classification problems with distribution shifts existing among the structure of different citation networks. 
\paragraph{Dataset.}

The Citation benchmark we use comprises three real-world paper citation networks~\cite{shen2020adversarial}, namely ACMv9, Citationv1, and DBLPv7, where ACMv9 network consists of papers published after 2010 from ACM, Citationv1 network is collected from Microsoft Academic Graph with papers published before 2008, and DBLPv7 network contains papers published between 2004 and 2008 on DBLP. Each node corresponds to a paper in the citation networks, and edges signify the citation relationships.  Each node belongs to some of the five categories:  Computer Vision, Databases, Networking, Information Security, and Artificial Intelligence. Table \ref{tab:dataset_citation} has listed the detailed statistics of the citation network dataset.
For simplicity, we use A, C, and D to represent ACMv9, Citationv1, and DBLPv7, respectively. More details regarding experiments can be found in the appendix.

\paragraph{Parameter settings.}
1) For low-weight edge dropping, without loss of generality, the sampling strategies use the \textit{threshold cutoff} (Eq.~\eqref{eq:threshold_cutoff_dorpedge}), and the hyper-parameter $\alpha,\rho$ is adjusted via the grid search strategy.
2) For spectral clustering-based edge adding,
The similarity matrix in spectral cluster employs the 
Gaussian kernel function RBF, i.e.
\begin{align}
    \S_{ij}=\exp(-\frac{\|\mbf x_i-\mbf x_j\|_2^2}{2\zeta^2}),
\end{align}
where $\zeta$ is a hyper-parameter that controls the distance flatness.
 The default number of clusters $\mcal K$ is set to $100$, which also can be adjusted during training. 
 We use the \textit{threshold cutoff} (Eq.~\eqref{eq:threshold_cutoff_dropadd}) strategy to generate the sampling probability and use the simple graph merge method (Eq. ~\eqref{eq:simple_new_graph}) to generate the new graph augmentation, which corresponds to the "low weight + clustering" in our experiments.
 The hyper-parameter $\beta, \epsilon, \rho$ is selected by the grid search strategy. 3) For model optimization, we use Adam optimizer with a learning rate of 0.001 for all experiments.

\paragraph{Evalutaion.} We adopt the leave-one-domain-out evaluation protocol in alignment with previous works\cite{xu2021fourier}, i.e. select one domain as the test domain and train the model on all other domains. Thus, three domain generalization tasks can be constructed: AC $\rightarrow$ D, AD $\rightarrow$ C, and CD $\rightarrow$ A.

As the citation networks used are multi-class data, we utilize Micro-F1 and Macro-F1 as the evaluation metrics to showcase the classification performance.
Micro-F1 and Macro-F1 are two  various average forms of the F1 score, a measure that emerges to take balance between the two significant but contradictory measures: precision and recall under the binary classification problem.

The formula is as follows:
\begin{align}
    \text{F1}=\frac{2\times \text{Precision} \times \text{Recall}}{\text{Precision} + \text{Recall}}=\frac{2\times TP}{2\times TP+FP+FN},
    \label{eq:f1_score}
\end{align}
where, $TP$, $FP$, and $FN$ are the numbers of true positive samples, false positive samples and false negative samples. In multi-class classification problems, which also can be seen as a combination of multiple binary classifications, the Micro-F1 calculates the overall F1 score of these binary classifications by first averaging each item $\overline{TP}$, $\overline{FP}$, and $\overline{FN}$ in the above Eq.~\eqref{eq:f1_score}, while the Macro-F1 score averages the F1 score obtained from each binary classification problem. 
To put it simply, the Micro-F1 score emphasizes the performance of the classifier on individual samples; whereas the Macro-F1 score accentuates the classifier performance on each class irrespective of the number of instances it has.

\subsection{Comparison with SOTA }
In this part, we combine our proposed augmentation methods with different graph neural network frameworks to evaluate the effectiveness of our methods. We adopt the proposed augmentation techniques in GCN and GIN to compare with the classical models: GCN~\cite{kipf2017semi}, GIN~\cite{xu2018powerful}, GAT~\cite{velivckovicgraph}, SGC~\cite{wu2019simplifying}, and the results are shown in Table \ref{tab:citation_benchmark}. According to the results, we have the observation both GCN and GIN with the proposed augmentation technique exceed largely their counterparts and the other two frameworks. The significant performance improvement demonstrates the effectiveness of our augmented technique in tackling the graph domain generalization problems. This can be attributed to two aspects: (i) the augmentation technique increases the number of training samples; (ii) the proposed augmentation technique generates more samples out of the domain.

\subsection{Comparison with Various Augmentations }
In this experiment, we intend to evaluate the power of our proposed graph augmentation technique with other graph augmentation methods. Since we have not found related works tailored for cross-graph node classification with OOD structure, we adopt several typical graph augmentations, including random \textit{Edge Dropping}, random \textit{Edge Adding}, random \textit{Feature Masking} and random \textit{Feature Dropout}.
\vspace{-2.5mm}
\begin{itemize}
    \item \textbf{Edge Dropping.} The edge dropping is a widely used simple augmentation method that only randomly removes the existing edges in the graph with a certain proportion. 
    
    \item \textbf{Edge Adding.} The edge adding is to randomly add extra edges to the graph with a certain proportion.

    \item \textbf{Feature Masking.} Feature masking aims to set several columns of the feature attribute matrix to be zero randomly and use the remaining attribute for model learning.

    \item \textbf{Feature Dropout.} Feature dropout aims to set a part of the feature attribute to be zero randomly and use the remaining attribute for model learning.
\end{itemize}
\vspace{-2.5mm}
Notably, node-level augmentations cannot be used as baselines, because our task is full-supervised node classification. 

\paragraph{Results.} We report the comparisons with the above typical augmentations upon the GIN backbone in Table \ref{tab:tab:augmentation_GIN_all}. 
From the results, we can know that
our augmentation techniques: low weight dropping and low weight dropping + spectral clustering consistently outperform the considered augmentation methods in all three tasks, regarding the classification Micro F1 score and Macro F1 score, suggesting that our methods achieve start-of-the-art  performance among conventional augmentations.
The results also reveal that in addition to improving the diversity and quantity of training samples, the proposed method is capable of boosting the cross-graph node classification performance of the graph model by investigating domain-invariant information.

\subsection{Ablation Studies}
\paragraph{Comparison with Random Edge Dropping.} We construct ablation studies on the effect of different drop rates in the low-weight edge-dropping augmentation and meanwhile make comparisons with the most related augmentation method: random edge-dropping. The results are shown in Figure \ref{fig:all_acc_gin}.  
The results suggest that the low-wight edge dropping outperforms random edge dropping consistently. When the ratio of dropped edges is within a certain range, the performance increases as the ratio of dropped edges increases, indicating some edges in the original graph involve noise that may hurt the generalization of the learned GNN. Further, the low-weight edge dropping achieves better than random edge dropping  with respect to noise edge removal.

\paragraph{Effect of Different Components.}
We conduct ablation studies to evaluate the performance of the two proposed methods. The results are shown in Table ~\ref{tab:ab_study_edge}. From the table, we have the following observation: \textbf{Observation} (1): Joining low-weight edge dropping and spectral clustering-based edge adding achieves the best performance, suggesting the two strategies can work in tandem. \textbf{Observation} (2): Adding the clustering edges into the original graph without low weight dropping leads to poor performance, implying that the edges in the original graph may contain some edge that will destroy the invariant structure constructed by spectral clustering.
Moreover, when we only use the clustering edge as the graph structure, namely, removing all edges in the original graph, we get a competitive performance. This also suggests that the clustering edge can really improve the GNN's ability to perceive the global node feature distribution.


\section{Conclusion and Future Work}
This paper introduces a simple and effective graph augmentation strategy for cross-graph node classification with OOD structure. 
There are still several directions that are worth exploring in the future: 1) the edge-dropping weight can be considered the more comprehensive method that can measure the significance of each edge. 2) The proposed augmentation can be extended to test time training.

\bibliography{0_main}

\clearpage
\onecolumn
\appendix
    \begin{center}
    \Large
    \textbf{Appendix}
     \\[20pt]
    \end{center}
   
\section{Details of experiments}
\subsection{Dataset}
\begin{figure*}[th]
    \centering
    \includegraphics[width=0.88\linewidth]{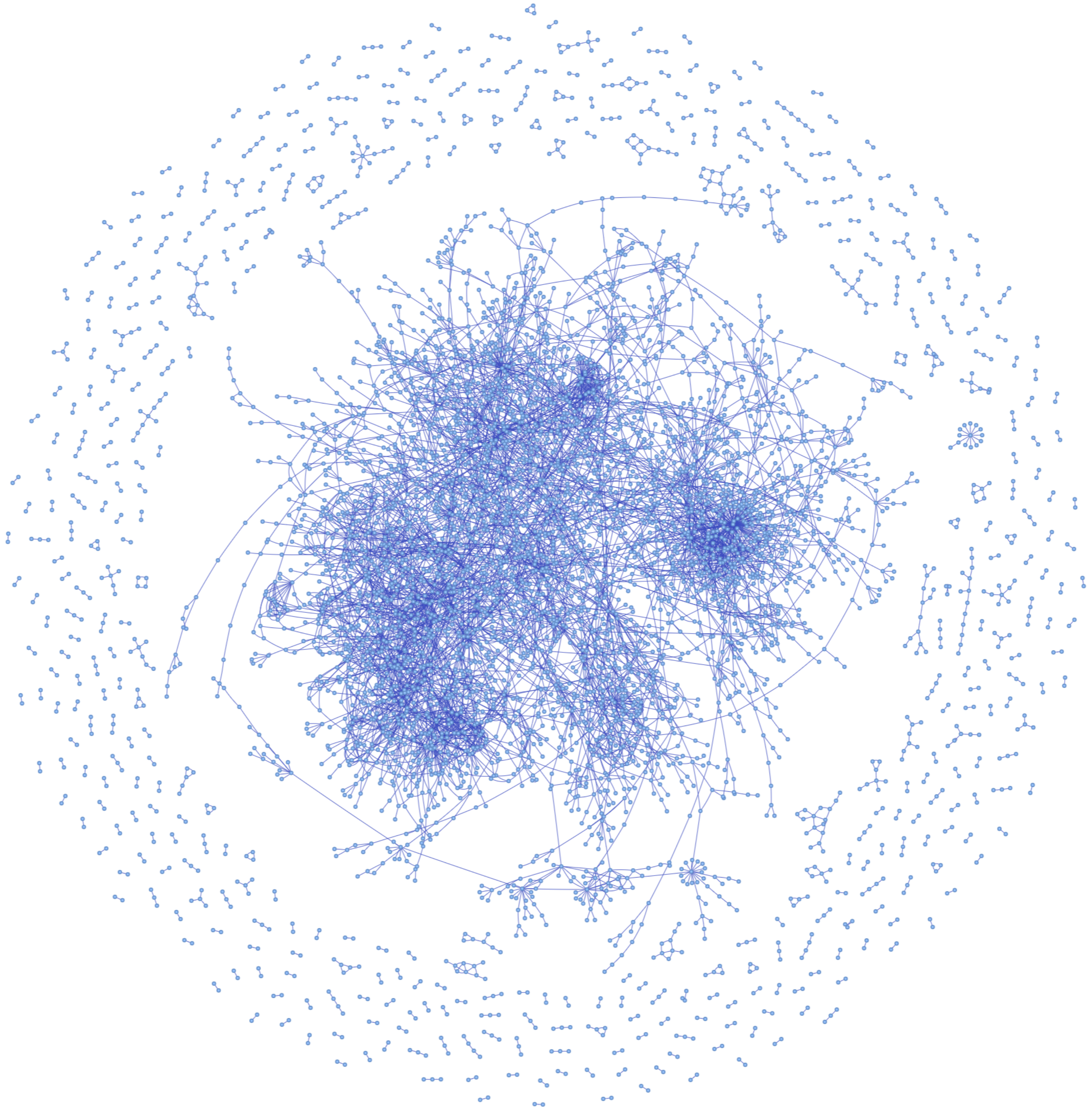}
     \caption{The visualization of the DBLP dataset}
    \label{fig:dblp_all}
\end{figure*}

The node classification benchmark with structured OOD for real-world citation networks is constructed based on the data in ArnetMiner\cite{tang2008arnetminer}, with papers published in various time ranges and collected from various sources. The Citation benchmark comprises three real-world paper citation networks, namely ACMv9, Citationv1, and DBLPv7, where ACMv9 network consists of papers published after 2010 from ACM, Citationv1 network is collected from Microsoft Academic Graph with papers published before 2008, and DBLPv7 network contains papers published between 2004 and 2008 on DBLP. In the citation networks, each node corresponds to a paper and edges signify the citation relationships. As for the node attributes, we generate bag-of-words vectors utilizing all the keywords from the paper titles. Each node belongs to some of the five categories:  Computer Vision, Databases, Networking, Information Security, and Artificial Intelligence. Table \ref{tab:dataset_citation} has listed the detailed statistics of the citation network dataset.
and Figure \ref{fig:dblp_all} is the visualization of the DBLP dataset.
In practical, we use the data from a public  (\href{https://github.com/shenxiaocam/ACDNE/tree/master/ACDNE_codes/input}{GitHub repository})\cite{shen2020adversarial}.
\begin{figure*}[th]
    \centering
    \begin{minipage}[s]{0.4\linewidth}
    \centering
    \includegraphics[width=1.0\linewidth]{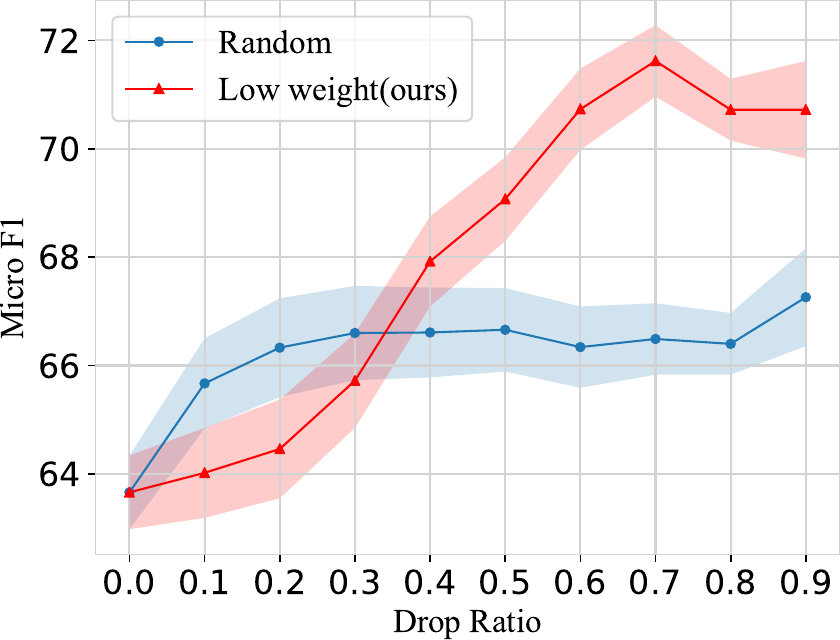}
    \end{minipage}
    \begin{minipage}[s]{0.4\linewidth}
    \centering
    \includegraphics[width=1.0\linewidth]{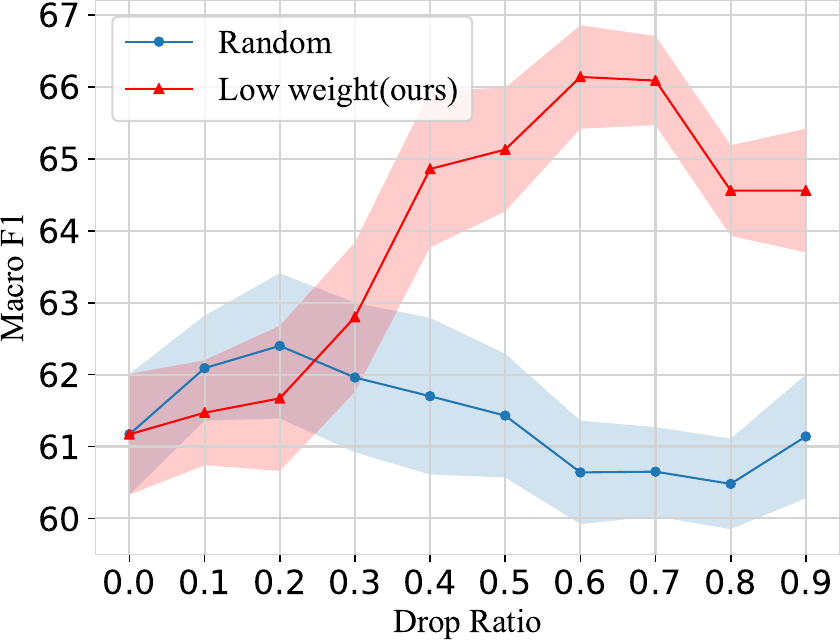}
    \end{minipage}

     \caption{The comparison between random edge-dropping and low weight dropping (backbone: GCN, AD $\rightarrow$ C)}
    \label{fig:all_acc_gcn}
\end{figure*}

\subsection{Additional Experiments}
Figure \ref{fig:all_acc_gcn} shows the results concerning the comparison between our  low weight dropping and random edge-dropping.
\end{document}